# CathAI: Fully Automated Interpretation of Coronary Angiograms Using Neural Networks

Robert Avram[1], Jeffrey E. Olgin[1,2], Alvin Wan[3], Zeeshan Ahmed[4], Louis Verreault-Julien[4], Sean Abreau[1], Derek Wan[3], Joseph E. Gonzalez[3], Derek Y. So[4], Krishan Soni[1], Geoffrey H. Tison[1,2,3,5]

1      Division of Cardiology, Department of Medicine, University of California, San Francisco, Cardiology (San Francisco, CA, United States), 505 Parnassus Avenue, San Francisco, CA, 94143, United States of America

2      Cardiovascular Research Institute, University of California, San Francisco, CA, 94143, United States of America

3      Department of Electrical Engineering and Computer Science, RISE Lab, University of California, Berkeley (Berkeley, CA, United States), Soda Hall, Berkeley California 94720-1770, United States of America

4      Division of Cardiology, Department of Medicine, University of Ottawa Heart Institute, University of Ottawa (Ottawa, ON, Canada), 40 Ruskin Street, Ottawa, Ontario, K1Y 4W7, Canada

5      Bakar Computational Health Sciences Institute, University of California, San Francisco, 94158, United States of America

**Corresponding Author:**

Geoffrey H. Tison MD MPH

Division of Cardiology, Department of Medicine

Bakar Computational Health Sciences Institute

University of California, San Francisco

San Francisco, CA 94158

Geoff.Tison@ucsf.edu

**Abbreviations:**

- **CHD:** Coronary Heart Disease
- **LCA**: Left Coronary Artery
- **RCA**: Right Coronary Artery
- **AUC:** Area Under the Receiver Operating Characteristic curve
- **IoU**: Intersection-over-union
- **PR-AUC:** Area Under the Curve for Precision and Recall Curve
- **QCA:** Quantitative Coronary Angiography
- **RAO:** Right anterior oblique
- **LAO:** Left anterior oblique
- **LOVI:** Layer Ordered Visualization of Information (explainability method)

**Abstract:** Coronary heart disease (CHD) is the leading cause of adult death in the United States[1] and worldwide[2], and for which coronary angiography is the primary gateway for diagnosis and clinical management decisions[3]. The standard-of-care for interpretation of coronary angiograms, including assessing CHD severity by quantifying coronary artery stenosis, depends upon ad-hoc visual assessment by the physician operator. However, ad-hoc visual interpretation of angiograms is poorly reproducible, highly variable and bias prone[4–8]. Here we show for the first time that fully-automated angiogram interpretation to estimate coronary artery stenosis is possible using a sequence of deep neural network algorithms. The algorithmic pipeline—called CathAI—achieves state-of-the art performance across the sequence of tasks required to accomplish automated interpretation of unselected, real-world angiograms. CathAI (Algorithms 1-2) demonstrated positive predictive value, sensitivity and F1 score of ≥90% to identify the projection angle overall and ≥93% for left/right coronary artery angiogram detection, the primary anatomic structures of interest. To predict obstructive coronary artery stenosis (≥70% stenosis), CathAI (Algorithm 4) exhibited an area under the receiver operating characteristic curve (AUC) of 0.862 (95% CI: 0.843-0.880) at the artery-level. When externally validated in a healthcare system in another country, CathAI AUC was 0.869 (95% CI: 0.830-0.907) to predict obstructive coronary artery stenosis. Our results demonstrate that multiple purpose-built neural networks can function in sequence to accomplish the complex series of tasks required for automated analysis of real-world angiograms. Deployment of CathAI may serve to increase standardization and reproducibility in coronary stenosis assessment, while providing a robust foundation to accomplish future tasks for algorithmic angiographic interpretation.

**Introduction:**

CHD results from atherosclerotic plaques narrowing the coronary arteries which can limit blood flow to cardiac tissues and ultimately lead to heart attacks[2]. Coronary angiography is the gold-standard minimally-invasive catheter-based procedure that provides assessment of coronary artery narrowing[3], or stenosis, and serves as the primary gateway for CHD treatment decisions. More than >1 million coronary angiograms are performed yearly in the United States alone[9]. The decision to treat CHD either through revascularization with coronary stents or bypass surgery, or CHD medications alone, relies upon angiography to identify coronary artery stenoses greater than 70% in severity[3].

The current standard-of-care ad-hoc visual estimation of coronary stenosis severity[10,11] has not changed in over 70 years and suffers from high inter-observer variability, operator bias and poor reproducibility[4–8]. Variability in visual stenosis assessment ranges from 15 to 45%[5,12–15], and is further exacerbated by substantial heterogeneity in operator experience: nearly 40% of cardiologists in the United States perform <50 angiograms yearly, which is considered low-volume[16]. Variability in coronary stenosis assessment has significant clinical implications, likely contributing to inappropriate coronary artery bypass surgery in 17% of patients and stent usage in 10% of patients[5]. A more standardized, reproducible approach to angiogram interpretation and coronary stenosis assessment therefore has critical clinical importance.

Existing methodologies designed to assist with quantifying angiographic coronary stenosis severity are not fully automated, require significant operator input and consequently are rarely used clinically. Quantitative coronary angiography (QCA) is the primary existing approach which requires manual selection of an optimal frame within the angiogram video, manual identification of a reference object (usually the guide catheter), and manual tracing of the vessel wall[5,6,17]. This manual input, required at multiple steps, is time-consuming and relegates QCA primarily to research applications[17]. Moreover, variability in QCA measurements range from 10-30%[15,18]. To date, full automation of coronary angiography interpretation has not been achieved likely due to the numerous complex sequence of component tasks, which presently is accomplished only through the expertise of highly sub-specialized physicians. The barriers to achieving automated angiogram analysis are varied and include frequent non-standard angiographic projections due to anatomic variation, multiple objects of interest that move throughout a video, variable contrast opacification of coronary arteries, plaque eccentricity, coronary artery overlap and "foreshortening" (the consequence of 2D visualization of 3D structures), and the need to integrate stenosis severity estimates from multiple frames within a video, and multiple videos that visualize the artery within a complete angiographic study to best determine an artery's percentage of stenosis[19].

To automate the process of angiogram interpretation and coronary artery stenosis estimation, we developed an algorithmic pipeline called CathAI consisting of multiple deep neural network algorithms which sequentially accomplish the necessary tasks for automated assessment of coronary stenosis severity. The sequence of tasks includes: classification of angiographic projection angle, anatomic angiographic structure identification (including left and right coronary arteries), localization of objects including coronary artery segments and stenosis, and determination of coronary stenosis severity. Importantly, our algorithmic pipeline also provides a broad foundation upon which to automate future downstream tasks in coronary angiogram interpretation.

During a standard coronary angiogram, catheters are inserted via the femoral or radial arteries and maneuvered through the aorta to canalize the coronary arteries. Fluoroscopic X-ray videos are acquired from different projection angles to visualize the coronary artery lumen during injection of iodine contrast from the catheter into a coronary artery. Each complete angiographic study consists of multiple videos that capture complementary projection angles to achieve optimal three-dimensional visualization of coronary arteries. After assessment of a stenosis in multiple projections, the performing cardiologist makes a visual estimate of the most severe percentage of narrowing from that stenosis (ranging from 0-100%) which is then described the procedural report[11]. Often, during the same procedure, coronary angioplasty and/or stents (wire meshes) are used to open severe coronary stenoses. For those deemed to have multiple severe stenoses or stenoses at critical locations not ideal for stents, bypass surgery would be recommended based on the angiogram findings.

**Overview of CathAI: A pipeline for automated angiographic interpretation**

CathAI consists of a sequence of 4 neural network algorithms organized in a pipeline (Figure 1a), with angiographic images "flowing" from one algorithm to the next, to accomplish the following tasks (see Methods):

a) Classification of angiographic projection angle (Algorithm 1)

b) Classification of an angiographic image's primary anatomic structure (Algorithm 2)

c) Localization of relevant objects within an angiogram, including coronary artery sub-segments and coronary stenoses (Algorithm 3 a/b)

d) Prediction of the coronary stenosis severity (as a percentage of artery narrowing) within an image containing a coronary artery segment (Algorithm 4).

Each algorithm was developed using training and test (and if appropriate, development) datasets tailored to that algorithm and task and consisting of non-overlapping patients (Methods). The Full Dataset, from which all subsequent angiogram datasets were derived, consisted of 13,843 de-identified coronary angiographic studies (195,195 total videos) from 11,972 patients aged ≥18 years from the University of California, San Francisco (UCSF), between April 1, 2008 and December 31, 2019 (Extended Figure 1). Mean age was 63.5 years (SD 13.7). Up to 8 frames were extracted from each Full Dataset video yielding 1,418,297 extracted Full Dataset images (Methods).

**Figure 1a.**

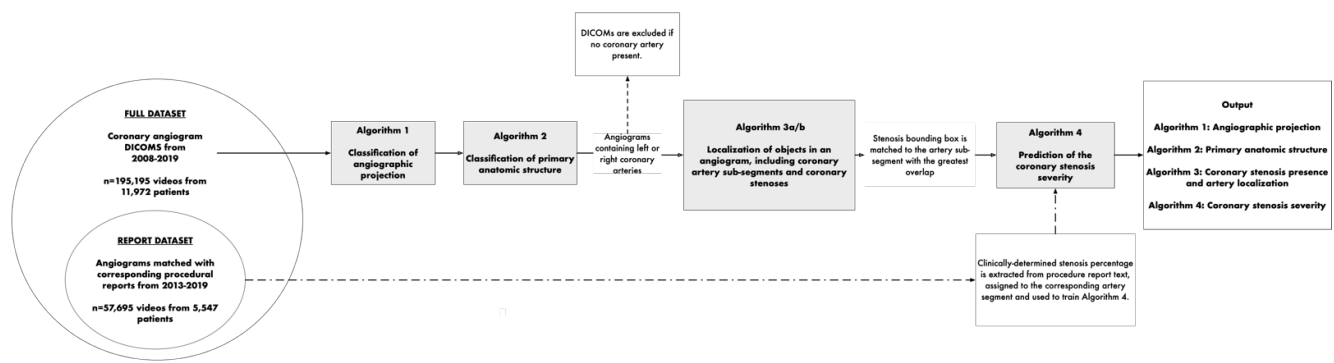

**Figure 1b.**

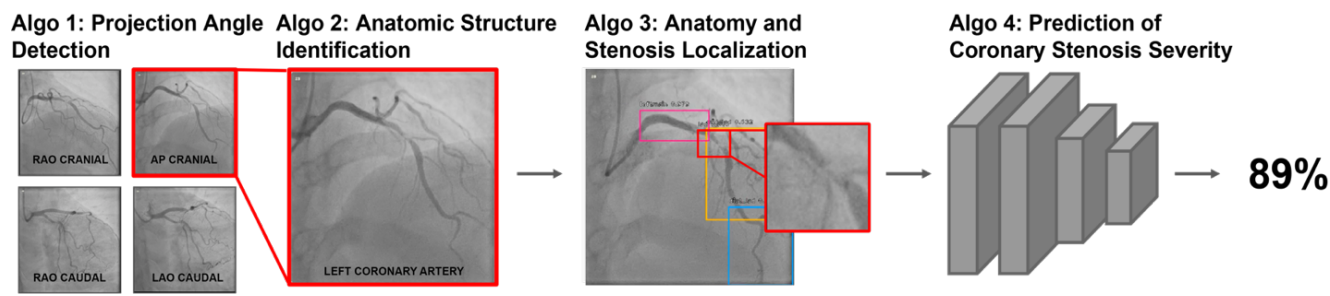

**Figure 1. Overview of CathAI. a. Overview of the CathAI pipeline for automated angiogram interpretation.** Diagram of the individual algorithms and primary datasets used in CathAI. Angiogram videos flow from one algorithm to the next, and CathAI provides a set of outputs, as shown, for each angiogram video. Algorithm 4 was trained on the Report Dataset, which was a subset of the Full Dataset that had corresponding digital procedural reports. Clinically-determined stenosis values were extracted from these reports and used to train Algorithm 4 (dash-dotted line). **b. Application of CathAI to an example coronary angiogram**. An example angiogram image of a left anterior descending artery with severe stenosis (in the proximal to mid segment) is shown progressing through each algorithm of the CathAI pipeline. Algorithm 1 identifies the angiographic projection, Algorithm 2 identifies that the LCA is present (left). Algorithm 3 then localizes objects in the image by predicting bounding boxes around objects, including coronary artery segments and stenoses. The stenosis is shown in zoom (middle). The bounding boxes are then used to automatically crop images around coronary artery stenoses to the nearest of three image sizes (aspect ratios) to enable input into Algorithm 4. Algorithm 4 predicts the maximal stenosis present in the image (AI-stenosis).

**CathAI performance to classify angiographic projection angle, anatomic structures, and object localization (Algorithms 1-3)**

Algorithm 1 identified the projection angle used in an angiographic image (Figure 1b). Extracted Full Dataset images were divided into training (990,082), development (128,590) and test datasets (299,625). The left-right and cranio-caudal projection angles recorded in metadata for each video were grouped into 12 distinct categories (Supplemental Table 1), and used to train Algorithm 1. In the hold-out test dataset, not previously used during training, Algorithm 1's frequency-weighted precision=0.90, sensitivity=0.90 and F1 score=0.90 (Extended Figure 2 & 3a). Performance was worse in the less commonly used antero-posterior and right anterior oblique (RAO)-lateral projections, and on the heterogeneous "other" class, defined as images that were not members of other listed classes. Once trained, Algorithm 1 was applied to all 1,418,297 extracted Full Dataset images. The most common prediction across extracted the frames of each video was assigned as the angiographic projection of that video.

Algorithm 2 identified the primary anatomic structure present in an angiographic video (Figure 1b). This enabled CathAI to focus subsequent analysis on videos containing coronary arteries, since angiograms commonly capture non-cardiac anatomic structures, such as the femoral or radial artery. Training data for Algorithm 2 was generated by manually classifying 14,366 angiographic images randomly selected from the extracted Full Dataset images (divided into 9,887 training, 1,504 development and 2,975 test datasets) into 11 classes describing the primary anatomic structure in the image (Supplemental Table 2). Image-level predictions were averaged to determine a video's primary anatomic structure. In the test dataset, Algorithm 2's frequency-weighted average precision=0.89, sensitivity=0.89 and F1 score=0.89 (Table 1). Since CathAI relied on Algorithm 2 to identify left and right coronary arteries, we generated more training data for these classes resulting in F1 scores of 0.95 and 0.93, respectively. F1 scores and model performance varied for other anatomic classes, but in general classes with lesser frames had lower performance (Extended Figure 3b), suggesting a possibility to improve performance if more labeled data were available. Once trained, Algorithm 2 was deployed on all contrast-containing extracted Full Dataset images to identify videos primarily containing the left and right coronary artery to flow into Algorithm 3; we identified 82,150 videos of the left coronary artery (LCA; 10,637 patients, 11,959 studies) and 32,318 videos of the right coronary artery (RCA; 9,599 patients, 10,535 studies). Other videos were excluded from further processing by CathAI.

Algorithm 3 localized relevant objects in angiograms, including coronary artery sub-segments and stenoses, by predicting surrounding bounding boxes (Figure 1b, Supplemental Video). To train Algorithm 3, we expert-labeled 2,338 peak-contrast frames from a random selection of Full Dataset videos (1,523 frames of the left and 815 frames of the RCA) by placing bounding boxes around the 11 dominant coronary artery segments (per SYNTAX[20]), coronary stenoses and other objects (Supplemental Table 3). We trained two versions of Algorithm 3. We first trained a general Algorithm 3a that was trained on all images and accepted both LCA and RCA images as input. To examine the benefit of training a dedicated algorithm for specific projection, we trained a specialized Algorithm 3b that accepts only RCA images in the left anterior oblique (LAO) projection as input, since the RCA is anatomically distinct and the LAO projection had the greatest total number of annotated images in our dataset (Methods). We then applied a post-hoc heuristic to exclude certain Algorithm 3a/b-predicted artery segments from angiographic projections (predicted by Algorithm 1) in which a-priori they are recognized not to be visible or are known to be foreshortened (Supplemental Table 4; Methods).

To assess the performance of Algorithm 3a/3b, we calculated the area of intersection over the area of union (IoU) between predicted bounding-box coordinates and the expert-annotated bounding-box coordinates of objects in each class in the test dataset. An IoU≥0.5 signifies at least 50% overlap between the predicted and true bounding-boxes, which we considered a true positive. We then measured the mean average precision (mAP) metric, which represents the ratio of true positives over true and false positives at different thresholds of IoU, for every class. A mAP value of 50% compares with state-of-the-art results for this type of task[21]. In the hold-out test dataset at the image-level, Algorithm 3a exhibited a 48.1% weighted average mAP (Extended Figure 4); average mAP was 37.0% for LCA segments, 42.8% for RCA segments, and 13.7% for stenosis. Algorithm 3b's weighted average mAP was 58.1%; average mAP was 54.5% for RCA segments and 26.0% for stenosis (Extended Figure 4). Once trained, Algorithms 3a/b were deployed on all images from videos primarily containing Left or Right Coronary arteries, as determined by Algorithm 2. The location of any identified coronary stenosis was assigned to the coronary artery sub-segment whose bounding box exhibited maximal overlap (by intersection over union) with a coronary stenosis bounding box (Figure 1b).

**Table 1. CathAI classification of the Anatomic Structure at the frame-level in the test dataset (Algorithm 2)**

| Anatomic Class | Precision | Recall | F1 Score | Number of Images |
|---|---|---|---|---|
| Left coronary artery | 0.97 | 0.94 | 0.95 | 1,055 |
| Right coronary artery | 0.93 | 0.93 | 0.93 | 632 |
| Bypass graft | 0.49 | 0.62 | 0.62 | 71 |
| Stenting procedure | 0.85 | 0.79 | 0.82 | 290 |
| Catheter | 0.78 | 0.91 | 0.84 | 512 |
| Pigtail Catheter | 0.69 | 0.55 | 0.55 | 44 |
| Ventriculography | 1.00 | 0.67 | 0.67 | 15 |
| Radial Artery | 0.55 | 0.63 | 0.63 | 11 |
| Femoral Artery | 0.95 | 0.97 | 0.97 | 286 |
| Aortography | 0.75 | 0.75 | 0.75 | 4 |
| Other | 0.44 | 0.44 | 0.44 | 55 |
| **Frequency-weighted average** | **0.89** | **0.89** | **0.89** | **2,975** |

Results are calculated on the hold-out Test Dataset for Algorithm 2.

**CathAI performance to predict stenosis severity (Algorithm 4)**

Once coronary artery segments and stenoses were identified by Algorithm 3a/b, the final task of the pipeline was to estimate the severity of identified coronary stenoses (Figure 1b). The primary goal was to estimate stenosis severity in a coronary artery segment based on all angiogram videos in a study that visualized that artery segment (which we call "artery-level"), mirroring standard clinical practice. To derive training data for stenosis severity, we obtained procedure reports containing the cardiologist-interpretation of UCSF coronary angiograms from January 1, 2013 to December 31, 2019. The maximal percentage and location of stenoses were extracted from these reports (called the REPORT-stenosis) and matched with their corresponding Full Dataset angiograms to derive the Report Dataset (Figure 1a, Methods). Mean age in the Report Dataset was 66.7±12.0 years. Algorithm 3a/b identified 4,328 Report Dataset angiograms with stenoses from 3,721 patients, totaling 46,168 videos (Extended Figure 1).

The Report Dataset was used to train Algorithm 4 to predict the maximal stenosis severity (among all visible stenoses) within input images, called AI-stenosis and expressed as a percentage from 0-100% (Methods). To predict binary "obstructive" coronary artery stenosis, defined as ≥70% stenosis[11], a threshold of 0.54 was chosen which optimized the F1 score. The mean stenosis prediction from all analyzed images in a video provided the video-level prediction, and the mean of video-level predictions from all videos that visualized an artery segment within a study provided the artery-level prediction. In the hold-out test dataset, Algorithm 4's AUC for obstructive stenosis was 0.862 (95% CI: 0.843-0.880) at the artery-level; AUC was 0.814 (95% CI: 0.797-0.831) at the video-level and 0.757 (95% CI: 0.749-0.765) at the image-level (Table 2; Figure 2a). Of those that had ≥70% stenosis according to the REPORT-stenosis, Algorithm 4 identified 74.5% correctly (95% CI: 70.0-78.4%; 260/349, Figure 2b); and of those that had <70% stenosis by REPORT-stenosis, Algorithm 4 identified 78.1% correctly (95% CI:76.1-80.1%; 1082/1385). Algorithm 4's operating threshold can be tuned to maximize sensitivity or specificity according to future target applications. However, we present several examples: when Algorithm 4's sensitivity to detect obstructive coronary stenosis was fixed at 80.0%, its specificity to detect obstructive stenosis was 74.1%; and when specificity was fixed at 80.0%, its sensitivity to detect obstructive stenosis was 71.6%. The mean absolute percentage difference between the AI-stenosis and report-stenosis was 17.9%±15.5% at the artery-level, 18.8%±15.8% at the video-level and 19.2%±15.1% at the frame level (Table 2). There was a significantly lower artery-level mean absolute percentage difference for the right coronary versus the LCA (16.4±15.0 vs 19.0±15.8; p<0.001, Table 2), at similar training dataset sizes—likely reflecting the RCA having less anatomic variation than the left. At the artery-level, there were medium-to-strong Pearson and intra-class correlations between the AI-stenosis and REPORT-stenoses values. Algorithm 4 tended to overestimate milder stenoses and underestimate more severe stenoses. There were only minor differences in performance between anatomic coronary artery segments, though mid vessels, generally had lower mean squared error and absolute difference compared to proximal or distal vessels.

**Figure 2.**

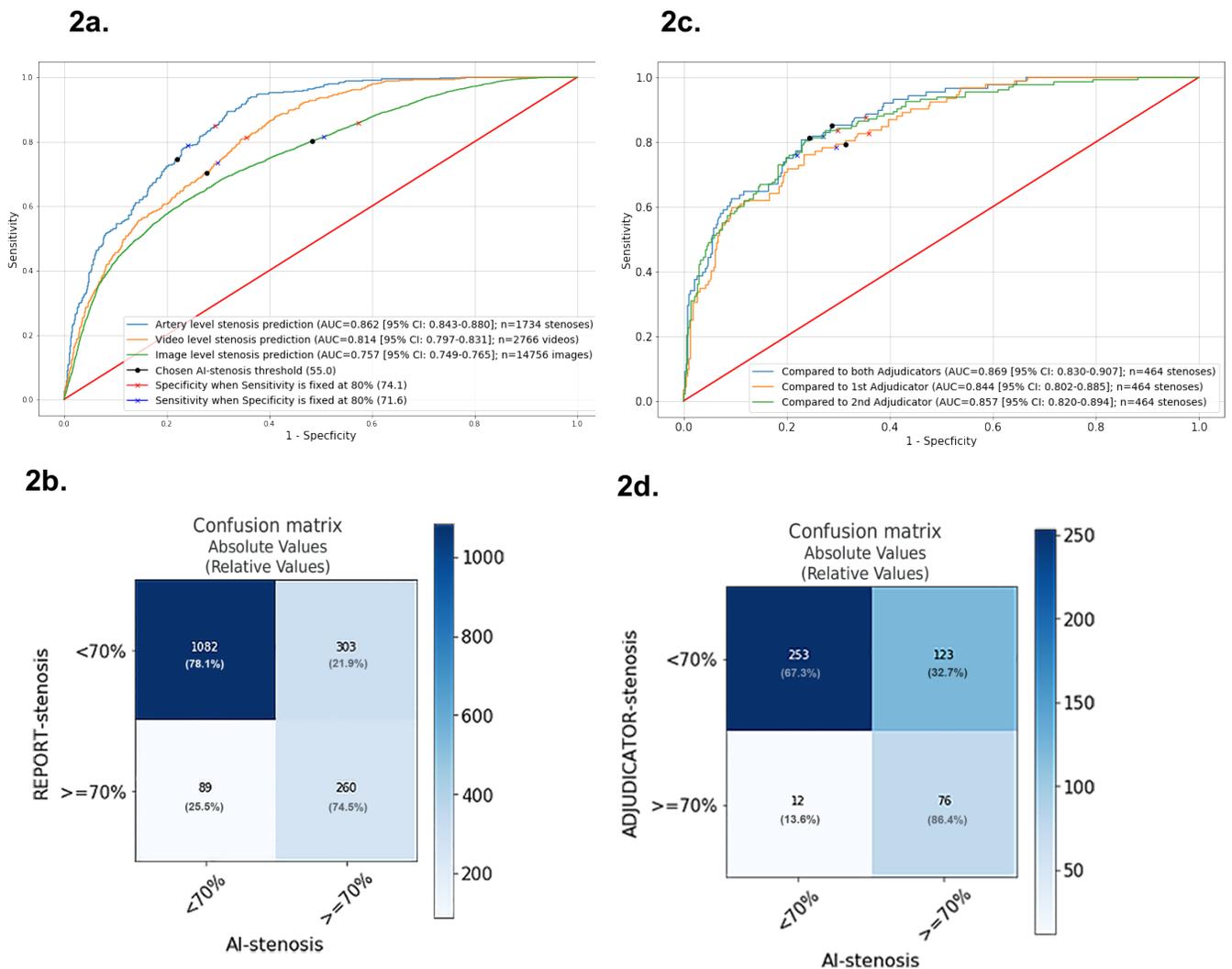

**Figure 2. CathAI performance to predict obstructive coronary artery stenosis. a. Algorithm 4a - Receiver Operating Characteristic Curves for CathAI prediction of obstructive (</≥ 70%) coronary stenosis in the test dataset.** Black dot: AI-stenosis threshold chosen to optimize F1 score in the artery-level dataset. Red cross: Specificity when sensitivity is fixed at 80%. Blue cross: Sensitivity when specificity is fixed at 80%. **b. Confusion matrix for Algorithm 4a prediction of obstructive stenosis vs. REPORT-stenosis in the test dataset. c. Algorithm 4a - Receiver Operating Characteristic Curves for CathAI prediction of obstructive (</≥ 70%) stenosis in an external validation dataset.** Black dot: AI-stenosis threshold chosen to optimize F1 score in the artery-level dataset. Red cross: Specificity when sensitivity is fixed at 80%. Blue cross: Sensitivity when specificity is fixed at 80%. **d. Confusion matrix for Algorithm 4a prediction of obstructive stenosis vs. expert adjudicators in an external validation dataset.** Abbreviations: AUC: area under the curve; CI: Confidence Interval; AI: Artificial Intelligence.

**Table 2. Performance of AI-stenosis (CathAI) versus REPORT-stenosis in the Test Dataset (Algorithm 4)**

|  | Number of REPORT-stenosis labels | AUC (95% CI) | Mean Absolute difference, % ± SD | r (95% CI) | ICC (95% CI) | Sensitivity (95% CI) * | Specificity (95% CI) * | PPV (95% CI) * | NPV (95% CI) * | MSE ± SD |
|---|---|---|---|---|---|---|---|---|---|---|
| **Artery level** | 1,734 | 0.862 (0.843-0.880) | 17.9 ± 15.5 | 0.74 (0.72-0.76) | 0.72 (0.60-0.84) | 78.1% (70.0-78.4) | 74.4% (76.1-80.1) | 46.1 (42.2-50.2) | 92.4 (90.9-93.6) | 561.99±850.01 |
| Left Coronary Artery | 976 | 0.835 (0.808-0.862) | 19.0±15.8 | 0.72 (0.68-0.74) | 0.69 (0.54-0.85) | 81.2 (67.5-86.2) | 70.6 (67.4-73.7) | 42.6 (38.0-47.1) | 93.3 (91.8-95.0) | 611.3±881.8 |
| Right Coronary Artery | 771 | 0.894 (0.869-0.919) | 16.4±15.0 | 0.77 (0.74-0.80) | 0.75 (0.65-0.87) | 74.7 (69.9-81.1) | 82.8 (80.0-85.6) | 50.1 (43.0-57.1) | 82.7 (80.1-85.6) | 494.9±798.8 |
| **Video level** | 2,766 | 0.814 (0.797-0.831) | 18.8±15.8 | 0.70 (0.68-0.72) | 0.66 (0.54-0.80) | 70.2 (66.7-73.5) | 72.4 (70.6-74.3) | 44.1 (41.2-47.0) | 88.7 (87.2-90.0) | 602.5±868.9 |
| **Frame level** | 14,756 | 0.757 (0.749-7.645) | 19.2±15.1 | 0.59 0.58-0.60) | 0.50 (0.35-0.70) | 80.1 (79.1-81.0) | 51.7 (50.7-52.6) | 52.9 (52.0-53.9) | 79.3 (78.4-80.2) | 594.97±855.24 |

**Abbreviations:** ICC: Intra-class correlation; *r*: Pearson correlation; AUC: Area Under the Receiver Operating Characteristic Curve; PPV: Positive Predictive Value; NPV: Negative Predictive Value; MSE: Mean Squared Error

*The threshold for determining sensitivity, specificity, PPV and NPV for AI-stenosis was 54%.

**External validation of CathAI in a different hospital**

To examine the external generalizability of CathAI, we obtained an external validation dataset of 464 angiogram videos from a quaternary care institution (University of Ottawa Heart Institute, Canada) for which stenosis severity was independently adjudicated for this purpose by two blinded experts from that institution. Inter-observer variability for stenosis severity was 15.7%±14.5%. For determination of obstructive stenosis (</≥70%), adjudicators disagreed on 16.8% of stenoses as being obstructive (n=78). The two adjudicators agreed on the localization of a stenosis within the same coronary artery segment in 91.4% of videos (n=424); in this subset CathAI (Algorithm 3) localized the stenosis in the same artery segment in 84.5% of the videos (n=360). We calculated the arithmetic mean percent stenosis from the two adjudicators to compare against CathAI's prediction of stenosis severity. Compared to this, CathAI's AUC for obstructive stenosis (≥70%) was 0.869 (95% CI: 0.830-0.907; Figure 2c); sensitivity was 86.4% and specificity was 67.3% (Figure 2d, using the same threshold from the derivation dataset of 0.54). The mean absolute percentage difference between AI-stenosis and the percent stenosis averaged from the two adjudicators was 18.02%±11.02%.

**Using Algorithm Explainability to understand CathAI performance**

To better understand what elements within angiograms contributed to CathAI predictions, we applied two explainability approaches to the fully-trained CathAI algorithms. This helps illuminate how the algorithms function and may provide additional decision-making context to clinicians. We applied the GradCAM technique[22] to highlight regions of the image most critical to an algorithm's prediction. Doing so demonstrated that Algorithm 2 often used similar image regions as a cardiologist to classify anatomic structures, such as the left anterior descending artery region to classify the image as containing the LCA (Figure 3a). For Algorithms 1 and 4, we derived Saliency maps using the Layer Ordered Visualization of Information (LOVI) method[23] which highlights individual pixels in the image that contribute most to each Algorithm's stenosis severity prediction (Figure 3b). Instead of localizing broad regions like GradCAM, LOVI highlights any number of individual pixels across an entire image, providing a more granular assessment of how disparate areas of the image assist Algorithm predictions. As shown in Figure 3b examples, highlighted pixels contributing to Algorithm 4's prediction were not only limited to stenosed artery segments, but also to normal artery segments and pixels immediately adjacent to the artery, suggesting that non-stenosed artery segments contribute to Algorithm 4's stenosis prediction.

**Figure 3**

**3a.**

**Original Angiogram Image**

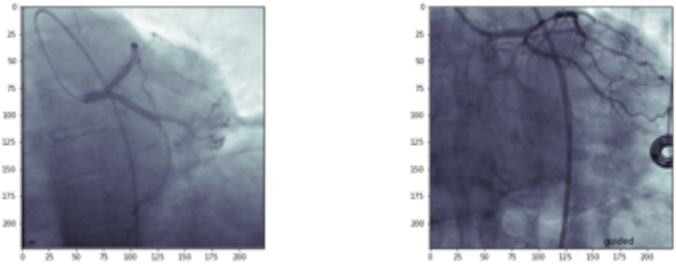

**GradCAM-highlighted Image**

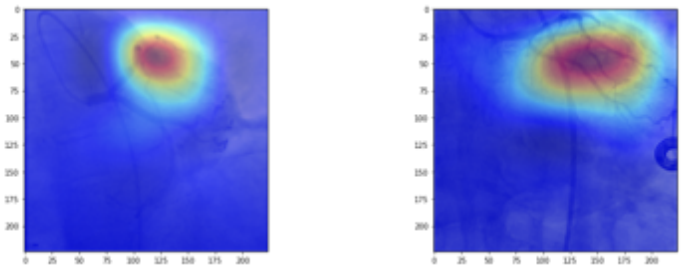

**3b.**

**Original Angiogram Image**   **LOVI Saliency Map**

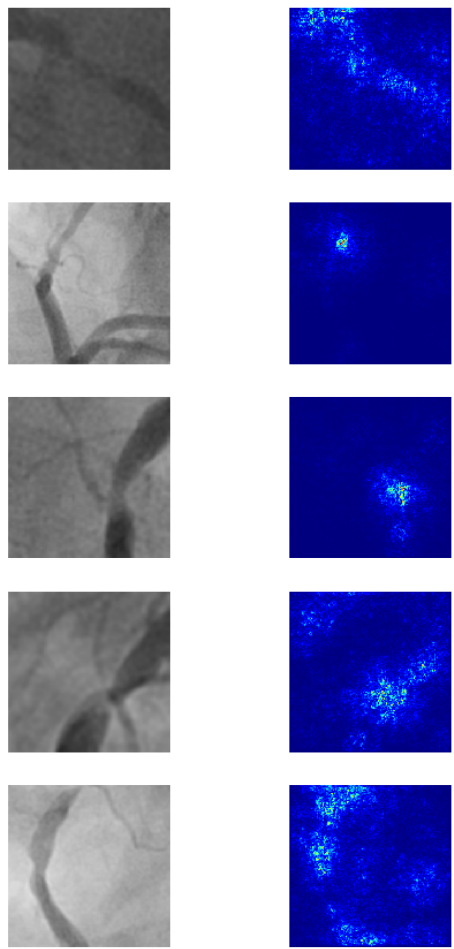

**Figure 3. Explainability methods applied to CathAI algorithms. a. GradCAM applied to Algorithm 2 classification of primary anatomic structure.** Two original angiogram images are shown (top), alongside corresponding images highlighted by GradCAM (bottom). GradCAM highlighted areas of the input angiogram image that Algorithm 2 used to identify anatomic structure in this example image of a LCA. Red areas denote higher importance to the algorithm prediction, whereas blue areas denote lower importance. **b. LOVI Saliency Maps of Algorithm 4 prediction of coronary stenosis severity.** Five original angiogram images are shown (left) which were input into Algorithm 4. Corresponding images with LOVI saliency maps are shown (right), with brighter pixels representing greater contribution to Algorithm 4's prediction. LOVI: Layer Ordered Visualization of Information.

**Discussion**

CathAI is a series of deep learning algorithms that provides the first fully-automated pipeline for coronary angiogram analysis. Because coronary angiography is the primary gateway for CHD clinical decision-making and the gold-standard for CHD diagnosis and quantification, CathAI has considerable potential for clinical impact by increasing standardization and decreasing variability over ad-hoc visual estimation. CathAI achieves state-of-the-art performance on each task required for automated angiogram interpretation on non-curated, real-world clinical angiograms. Beyond assessment of stenosis, this pipeline additionally can provide an algorithmic foundation for automation of future angiogram-relevant tasks—such as real-time recommendations for additional testing—without requiring manual screening for selected projections/views or artery tracing. CathAI predicts stenosis severity for any given angiogram with zero variability, and its deviation from human experts is well within, and often less than, the inter-observer variability that we observed and that has been previously reported[5,12–15]. Importantly, CathAI was generalizable, without any additional training, to real-world angiograms from a separate medical system in a different country. Automated angiographic analysis and stenosis assessment may increase standardization at one of the most critical stages in CHD clinical decision-making, assisting physicians and providing an objective standard to improve reproducibility in coronary stenosis assessment.

Automated analysis of angiograms has greater similarities to the perception-side of "self-driving car" technology than to standard radiologic analysis (i.e. X-rays), given angiograms' highly variable, operator-determined video acquisitions and the sequence of complex tasks required for analysis. Successful angiogram analysis requires the AI-system to be able to process any type of video encountered during real-world procedures, identify relevant images, then localize and quantify critical objects within these images with high accuracy. While prior efforts have attempted individual tasks such as frame extraction[24,25] or stenosis prediction from RCA images only[11,25–31], to our knowledge none have developed a fully automated pipeline, which we demonstrated with deployment and validation on unselected real-world angiograms. Across each of the sequential tasks critical for angiogram interpretation, CathAI achieved state-of-the-art performance[11,25–31]. While additional improvement in each of CathAI's individual algorithms will likely be needed prior to clinical deployment, our proof-of-concept work demonstrates that multiple purpose-built deep learning algorithms can overcome the barriers that to date have prevented full automation of angiogram analysis. Like self-driving car technology, however, much work remains before CathAI is "clinically ready" given the high-stakes nature of coronary stenosis assessment. Our work provides the framework for rapid future improvements such as by generating additional training data and refining algorithms for specific clinical tasks. The potential clinical impact is substantial, however, given angiography's central role in nearly all clinical decisions related to CHD.

The most immediate clinical implication of deploying CathAI would be to increase standardization in the assessment of coronary stenosis, which currently suffers from high variability and potential for operator bias. The variability in human expert assessment of stenosis of up to 45%[8,28,32,33] directly impacts use of life-saving revascularization therapies for CHD. A prior study that re-evaluated clinical angiograms by committee reported that the recommendation for coronary bypass surgery changed from "necessary/appropriate" to "uncertain/inappropriate" in 17-33% of cases, and in 10% of cases for stent placement[5]—leading to the suggestion for a second independent interpretation before revascularization. While highly impractical if independent interpretation is performed by a second cardiologist, CathAI could easily perform this function in an automated, reproducible manner. Since the CathAI pipeline requires ~2 seconds to analyze each video on current consumer-grade GPUs (NVIDIA GTX 1080 Ti), CathAI can provide near real-time predictions to physicians during the procedure while angiograms are performed to supplement physicians' own ad-hoc diagnosis. This could increase standardization in stenosis assessment and subsequent adherence to guideline-based CHD treatment. The CathAI pipeline also provides a foundation upon which a wide range of clinically relevant applications related to automated angiographic analysis can be trained with additional task-specific data. For example identifying poorly-visible objects like prior stents, collateral arteries or bypass graft sites, or real-time recommendation to acquire additional projection angles or diagnostic procedures like fractional flow reserve[34] or intravascular ultrasound.

    The current state-of-the-art for assisted coronary stenosis assessment primarily encompasses QCA[3], which still relies upon significant human input and exhibits significant variability. A study assessing 10 different QCA systems against a phantom stenosis gold-standard found absolute percentage differences of -26% to +29% in coronary stenosis assessments between systems[18]. For different individuals using the same QCA system an 11.2% absolute percentage difference was reported. For different core labs using the same QCA system, 5.5-10.4% difference was reported[15]. Notably,

on our post-hoc review of examples of angiograms where AI-stenosis and REPORT-stenosis were discordant, the REPORT-stenosis value was not always objectively more correct. Because the mean absolute difference between AI-stenosis and REPORT-stenosis of ~18% was so similar to human inter-observer variability[33] (which is also likely present in the clinically-generated Report dataset), CathAI performance may be substantially improved if re-trained using less variable training labels—such as with core lab-generated QCA estimates. Therefore, QCA or other stenosis assessment assistance methods could play an important role to provide large numbers of high-quality training labels with which to further improve CathAI. Similarly, training dedicated algorithms for certain higher-volume angiographic projections, similar to our Algorithm 3B, could decrease input variability into all models and improve overall performance.

The model explainability methods we performed highlight how individual algorithms in CathAI function to accomplish their tasks. GradCAM and LOVI suggest that the algorithms often focused on similar regions of the image as a human expert does to classify anatomy and predict stenosis severity. For stenosis assessment (Algorithm 4), LOVI showed that CathAI involved pixels not only at the narrowest part of the stenotic artery but also outside of the area of stenosis, similar to how human cardiologists use both normal and abnormal segments of the coronary artery to assess relative severity.

Our work has several limitations. Due to the resource-intensive nature of generating large numbers of cardiologist annotations for angiographic images, we selectively used purpose-generated annotations for some tasks to prioritize efforts for the primary classes of interest (Algorithm 2, 3a/b). This resulted in greater amounts of training data for targeted classes, and fewer for others. To achieve the most data to train CathAI for stenosis severity, we used clinically-generated REPORT-stenosis values which were available in large numbers. Though these were generated by sub-specialty trained interventional cardiologists at UCSF, they likely still exhibit the variability inherent in any clinically-generated label. In addition, the text-parsing method we used to extract the REPORT-stenosis from the clinical procedure report may have introduced errors in either the location of the stenosis or the degree of severity. Resultant variability in stenosis labels used for training, either from clinical variability or parsing, would be expected to adversely impact algorithm performance by biasing results to the null and decreasing the observed effect of association. Perhaps in part due to this variability in the stenosis training labels, our algorithm tended to underestimate severe stenoses and overestimate minor stenoses. This may also have been impacted, in part, by having fewer training examples of severe stenosis. Future algorithm improvements will likely come from using purpose-generated labels, such as from a core-lab using either protocol-guided visual assessment or QCA, or physiologic assessment such as FFR.

**Conclusion**

CathAI provides the first fully automated analysis pipeline for coronary angiograms. CathAI achieves state-of-the-art performance for each task required for interpretation of real-world angiograms, and provides a foundation for future tasks in automated angiographic interpretation. The automated stenosis assessment enabled by CathAI may serve to increase standardization and reproducibility in coronary stenosis assessment, one of the most critical junctures in CHD clinical decision making.

**Acknowledgements:** Dr. Avram is supported by the "Fonds de la recherche en santé du Québec » (Grant 274831). Dr. Olgin received funds from the NIH (U2CEB021881). Dr. Tison received support from the National Institutes of Health NHLBI K23HL135274. Dr. Soni has no relevant funding related to this manuscript. Alvin Wan is supported by the National Science Foundation Graduate Research Fellowship under Grant No. DGE 1752814. Dr. So is supported by a Mid-Career Investigator Award by the Heart and Stroke Foundation of Ontario. Dr. Gonzalez is funded by the NSF CISE Expeditions Award CCF-1730628 as well as gifts from Amazon Web Services, Ant Group, CapitalOne, Ericsson, Facebook, Futurewei, Google, Intel, Microsoft, Nvidia, Scotiabank, Splunk and VMware. The funders had no role in study design, data collection and analysis, decision to publish or preparation of the manuscript.

**Author Contributions:** R.A., G.H.T., J.E.O., contributed to the study design. R.A., K.S., D.Y.S. and G.H.T. contributed to data collection. R.A. and G.H.T. performed data cleaning and analysis, ran experiments and created tables and figures. R.A., J.E.O., A.W., Z.A., L.V.J., S.A., D.W., J.E.G., D.Y.S., K.S. and G.H.T. contribued to data interpretation and writing. G.H.T. supervised.

**Competing Interests:** Dr. Tison has received research grants from Myokardia and Jassen Pharmaceuticals. Dr. Olgin has received research funding from the National Institute of Health (U2CEB021881; U54HL143541; 75N91020C00039; R01CA134722),

# Methods

## Study participants and study datasets

The Full Dataset consisted of retrospective, de-identified coronary angiographic studies from patients 18 years or greater from the University of California, San Francisco (UCSF), between April 1, 2008 and December 31, 2019 (Extended Figure 1). Angiograms were acquired with Philips (Koninklijke Philips N.V., Amsterdam, Netherlands) and Siemens (Siemens Healthineers, Forchheim, Germany) systems. For every angiographic video with contrast injection, we automatically identified the frames within the video that likely contained peak-contrast to generate individual images to be used as inputs into Algorithm 1-2 by calculating the structural similarity index between the first "reference" frame of the video (which typically does not contain intra-arterial contrast) and all subsequent frames of each video. The similarity index increases if two images have similar pixel values and decreases with greater difference. The frame with the lowest similarity index from the reference frame was selected as most likely containing peak-contrast (e.g. when the artery is full of contrast). Up to 8 frames were then extracted from each video to train Algorithm 1-2 (below): we retained the reference frame, the peak-contrast frame and the 3 frames immediately preceding and following the peak-contrast frame. This is referred to as the "extracted" Full Dataset frames (n=1,418,297). All frames of a video were converted to images of dimension 512*512 pixels for algorithmic analysis.

**The Report Dataset**

Shortly after performing the procedure, interventional cardiologists typically interpret the angiogram using visual assessment, as per standard clinical practice, and describe the severity of coronary stenosis in the procedure report. This procedural report text was parsed (see below) to identify: any description of coronary artery stenosis, the maximal stenosis percentage (called the REPORT-stenosis) and its location in one of 11 coronary artery segments (Supplemental Table 3). We identified 9,782 coronary stenoses in artery segments (REPORT-stenoses) and identified in 1,766 non-stenosed complete vessels yielding a total of 10,088 non-stenosed vessel segments, derived from 84,153 images. Then we randomly sampled 10,000 images corresponding to healthy artery subsegments (Extended Figure 1). Metadata was extracted from each angiogram video including the procedure date, the primary (Right Anterior Oblique [RAO]/Left Anterior Oblique [LAO]) and secondary (cranio-caudal) angles of rotation, a unique exam identifier and a unique patient identifier.

**Text Parsing Methods**

The free text from the procedural report was first segmented using commas (",") or periods ("."). We then applied text parsing methods to identify distinct coronary segments (Supplemental Table 3). When a coronary segment was found, we identified any description of corresponding stenosis percentage by localizing "%" and the nearest one to three-digit number in that sentence. We initially searched using standard terms (such as "right coronary artery"), then expanded the keywords by manual review of the text, over multiple iterations, to include the most common abbreviations, alternate spellings and permutations (Such as "RCA"). Qualitative descriptions of obstructive CHD, such as "mild", "moderate" or "severe" disease were not extracted. Furthermore, we searched for keywords such as "thrombus", "obstruction" or "occlusion" in the report; when present in a coronary segment, 100% stenosis was assigned to that segment. For analysis, ostial and proximal coronary segments were merged in the 'proximal' class and ostial, proximal, middle, and distal left main arteries were merged under the 'left main' class. The most severe stenosis within any of the 11 segments was retained (Supplemental Table 3). We did not analyze stenoses in diagonals, marginals, septals, ramus, left posterior descending artery, left posterolateral or in bypass grafts.

**Human Subjects Research**

This study was reviewed by the University of California, San Francisco Institutional Review Board and need for informed consent was waived. The external validation was reviewed and approved by the University of Ottawa Institutional Review Board.

**Algorithm Development**

**Algorithm 1: Classification of angiographic projection angle**

Algorithm 1 accepted individual images (video frames) as its input and identified the angiographic projection angle used. The projection angle refers to the fluoroscopic angulation used to obtain the image, commonly described on two axes defined by LAO-RAO and cranial-caudal (LAO cranial, RAO caudal, etc). To do so we used the 8 images from the extracted Full Dataset and labeled these images using the primary and secondary angles extracted from each video's metadata, into 12 classes of angiographic projections (Supplemental Table 1). Angles were extracted as two continuous variables ranging between -180 and 180 degrees for the primary angle and -50 and 50 degrees for the secondary angle. We then split the Full Dataset comprising 1,418,297 images from 11,972 patients and 195,195 videos for identifying angiographic projection into Training/Development/Test sets (990,082 images in Training, 128,590 images in Development and 299,625 images in Test datasets).

Algorithm 1 architecture was based on Xception, which is a convolutional neural network that has achieved state-of-the-art performance at image recognition tasks[35]. It was initialized with 'ImageNet' weights, as commonly performed to initialize weights for faster algorithm convergence in image classification settings[36]; all layers were trainable. Images were augmented by random zoom (range=0.2) and shear rotate (range=0.2). The development dataset was used to iteratively compare algorithm performance and fine tune hyperparameters using grid search. We experimented with different architectures such as VGG-16, ResNet50 and InceptionNet but found no incremental benefit over Xception. The Test dataset was not used at all during training, and was only used to report final performance. The most common prediction across extracted the frames of each video was assigned as the angiographic projection of that video; ties were addressed by selecting the projection with the highest average probability across all frames.

**Algorithm 2: Classification of primary anatomic structure**

Algorithm 2 aimed to identify the primary anatomic structure present in an angiographic video (Supplemental Table 2), since it is common to capture videos containing non-cardiac anatomic structures such as the aorta or the femoral artery during a coronary angiography procedure. We randomly selected 14,366 images from the extracted Full Dataset, and a cardiologist labelled each image into one of 11 classes. This dataset was split into Training/Development/Testing datasets, containing 9,887 (70%), 1,504 (10%), and 2,975 (20%) images, respectively. We trained Algorithm 2 using the Xception architecture, initialized weights from trained Algorithm 1, and tuned hyper-parameters. Images were augmented by random zoom (range=0.2) and shear rotate (range=0.2). The primary anatomic structure of a video was derived from the mode prediction of all extracted Full Dataset frames comprising that video. Only videos that primarily contained right or left coronary arteries flowed to Algorithm 3 for subsequent CathAI analyses (Extended Figure 1).

**Algorithm 3: Localization of angiogram objects**

Algorithm 3 aimed to localize relevant objects within images of the left and right coronary arteries (the output of Algorithm 2). While Algorithm 3 was trained to localize multiple objects (Supplemental Table 3), the tasks most critical to the CathAI pipeline were to (i) identify coronary artery segments, (ii) identify stenoses (if present) and (iii) localize other relevant objects such as guidewires or sternotomy. To generate training data for Algorithm 3, we labeled 2,338 images of left and right coronary arteries both with and without stenosis. Only stenoses in the main epicardial vessels, not side branches such as diagonals or marginals, were labeled. In our final pipeline we trained two versions of Algorithm 3: Algorithm 3a was trained on all available LCA and RCA angiogram videos. Since the RCA in the LAO projection contained the greatest number of annotated images, we also trained a dedicated Algorithm 3b to demonstrate possible performance gains from focusing an algorithm on a specific artery/projection (RCA in LAO). To train Algorithms 3a/b, we split our labelled images for this task into two separate datasets: One containing left/right coronary arteries (2,338 images) and one containing RCA images in the LAO projection (450 images). Each dataset was subsequently split into 90% training (2104 and 405 images respectively) and 10% test (234 and 45 images respectively). Algorithm 3a trained on 2,338 images with 12,685 total distinct classes and was trained for 50 epochs. Algorithm 3b was trained on 450 images with 2,447 total distinct classes and was trained for 50 epochs. Deployed in the CathAI pipeline, Algorithm 3b served to decrease input variability for Algorithm 3a, which resulted in performance improvements for both algorithms. Future improvements may be achieved with additional dedicated algorithms.

Algorithms 3a/b employed the RetinaNet architecture and were trained using original hyperparameters[37]; a development dataset was not used. RetinaNet has achieved state-of-the-art performance for object localization such as the pedestrian detection for self-driving cars[21] and in medicine, was used to localize and classify pulmonary nodules in lung CT-scans[38]. For our task, Algorithms 3a/b output bounding box coordinates for any objects present in each input image. Because some artery segments in certain angiographic projections are known *a priori* to be foreshortened or not visible, we applied a post-hoc heuristic to exclude certain Algorithm 3a/b-predicted artery segments from angiographic projections (Supplemental Table 4). To assess Algorithm 3a/b performance, the predicted coordinates were compared with the ground-truth coordinates using the ratio of the area of intersection over the area of union (called Intersection-over-union [IoU]). An IoU≥0.5 between the predicted and annotated coordinates was considered a true positive. Next, we measured the mean average precision (mAP), which represents the ratio of true positives over true and false positives at different thresholds of IoU, for each class. A mAP of 50% compares with state-of-the-art results for this type of task.

**Algorithm 4: Predicting the Percentage of Coronary Artery Stenosis**

Algorithms 3a/b were run on the Report dataset to localize artery segments and stenoses. All frames that contained a stenosis bounding box with an IoU[39] ≥0.20 with a coronary artery segment bounding box were used. A stenosis was assigned to the artery segment, as identified by Algorithm 3, with greatest overlap by IoU. To derive train/test labels for Algorithm 4, we cross-matched stenoses found by Algorithm 3a/3b with the stenosis percentage found in the procedural report in corresponding artery segments . Matched procedural report values served as labels to train Algorithm 4 with input images cropped around stenosed artery segments according to Algorithm 3a/b bounding boxes. Non-matched stenoses were removed from our dataset. We also excluded all videos where an intra-coronary guidewire was present in more than 4 frames, since these videos likely represent stenting procedures which would alter the stenosis percentage within that video. We retained all videos from a study prior to the insertion of an intracoronary guidewire. Once a stenosis was identified, bounding box coordinates were expanded by 12 pixels in all dimensions, then cropped and resized to the nearest of three predetermined sizes: 256*256 pixels (aspect ratio no.1), 256*128 pixels (aspect ratio no.2) and 128*256 pixels (aspect ratio no.3). This was performed to maximize signal-to-noise (vessel-to-background) ratio, due to different vessel orientations and stenosis sizes. The "Report Dataset" used for Algorithm 4 consisted of 105,014 images (6,667 stenoses coming from 2,736 patients and 5,134 healthy vessel segments from 1,160 patients; Extended Figure 1). Since non-stenosed vessel segments tended to be longer than focal stenosis which may bias training, we cropped all non-stenosed segments randomly to a height and width, mirroring the distribution of stenosis image sizes within that coronary segment. This yielded similar vessel sizes between the stenosed and non-stenosed images for each vessel segment. Images were randomly split into 70% training, 10% development and 20% in testing datasets.

Algorithm 4 was based on a modified Xception architecture where the last layer (Softmax layer, used for classification) was replaced with an 'average pool' then dense layer with a linear activation function to enable prediction of stenosis severity as a continuous percentage value. Image metadata consisting of the coronary artery segment label and cropped aspect ratios were also added as inputs into the final layer of Algorithm 4, which improved performance. The algorithm output a percentage stenosis value between 0 and 100 for every input image representing the maximal stenoses in that coronary artery segment. Model weights were initialized using those from the trained Algorithm 1. Images were augmented by random flip (both horizontal and vertical), random contrast, gamma and brightness variations, random application of CLAHE (To improve contrast in images)[40]. The algorithms were trained to minimize the squared loss between the predicted (AI-stenosis) and the report-stenosis using the RADAM Lookahead[38] optimizer with an initial learning rate of 0.001, momentum of 0.9 and batch size of 12, trained for 50 epochs[39]. Training was halted when loss stopped improving for 8 consecutive epochs in the test dataset.

We examined various pre-processing approaches and sequences without improvement in algorithm performance. For Algorithm 4, each epoch was trained on images of one aspect ratio, with the next epoch training on another aspect ratio (copying all weights from the previous iteration), as performed previously for multi-size inputs[41]. This was iterated until convergence. We measured the algorithm performance on the complete test dataset, consisting of the three aspect ratios. We observed that the convergence of the multi-size input training was similar to other algorithms that used a fixed size for training.

We examined the characteristics of those patients who were determined to have obstructive AI-stenosis (</≥70%) that were either concordant (1,336) or discordant (398) with the REPORT-stenosis. We specifically refrain from using the terms "false positive/false negative" in this setting, since as we discussed, the REPORT-stenosis is subject to error and non-trivial variability. AI-stenosis was more likely to be discordant with REPORT-stenosis in older patients (62.7±13.2 vs 65.1±12.3, <0.001), in the LCA, the proximal RCA, distal RCA, the right posterolateral and the distal LAD.

For all algorithms, except Algorithm 3, data was split randomly for each algorithm into Training (70%), Development (10%) and Test (20%) datasets, each containing non-overlapping patients. The development dataset was used for algorithm tuning, when required. For Algorithm 3, dataset splits were Training (80%) and Test (20%); since we used original hyperparameters[21,42] and did not require algorithm tuning.

**External Validation**

For external validation, we obtained 1000 consecutive coronary angiogram videos performed at the University of Ottawa Heart Institute (UOHI) between July 1st 2020 and October 31st 2020, acquired with Philips systems (Koninklijke Philips N.V., Amsterdam, Netherlands). Algorithms 1, 2 and 3 were applied to each video to identify and localize stenoses, and Algorithm 4 predicted AI-stenosis. We then sampled up to 40 examples of angiogram videos per coronary artery segment to form our external validation dataset, identifying 464 coronary angiograms with distinct stenoses. Two board certified cardiologists at the UOHI, each with over 2000 coronary angiograms of experience as primary operators, adjudicated these videos in a blinded fashion by grading stenosis severity as a percentage between 0-100% and localizing the stenosis to a coronary artery sub-segment. Algorithm performance in this dataset was reported as the AUC of the AI-stenosis compared to each adjudicator, and to the average of both adjudicators. Since stenoses in this external validation dataset were only visualized in one video, there was no calculation of artery-level AI-stenosis performance. The binary threshold for obstructive AI-stenosis was the same (0.54) as used in the primary analysis. We also described the concordance between the localization of the stenosis as determined by Algorithm 3 and by the two adjudicators as well as the average difference between each adjudicator stenosis percentage.

**An alternative approach to predict stenosis severity (Algorithm 5-6)**

We developed an alternative approach to Algorithm 4 to estimate stenosis severity that we present here as a sensitivity analysis. This approach aimed to eliminate the contribution of background non-artery pixels to the stenosis prediction, corroborated Algorithm 4's ability to predict stenosis without artery-masking, and provided an alternative approach to stenosis prediction that we postulate may excel in some settings. This parallel approach, described in greater detail below, first used Algorithm 5 to segment the boundaries of the coronary artery within a cropped input image (e.g. the output of Algorithm 3). Algorithm 5's predicted mask was used to set all non-artery pixel values to 0 (called the "Segmented image"). Algorithm 6 then predicted the percentage of stenosis from these images.

**Algorithm 5-6: A Generative Adversarial Network for Vessel segmentation**

Algorithm 5 segments the coronary artery within images cropped around coronary artery stenoses (by Algorithm 3a/b). The goal of artery segmentation was to classify pixels within a coronary artery-containing image into 'artery' or 'not-artery' pixels (called "pixel-wise segmentation"). Isolating the artery from the background would minimize non-artery input into stenosis predictions. To generate a training dataset for Algorithm 5, a cardiologist traced the vessel contour of 160 images of stenoses and 40 images of non-stenosed coronary segments to generate ground-truth masks used to train and test Algorithm 5. Annotated Algorithm 5 data was then divided into 90% training and 10% test datasets.

To perform this segmentation task we used a Generative Adversarial Network which previously demonstrated state-of-the-art automatic retinal vessel segmentation using small datasets (less than 40 images for training)[42]. This network learns to generate a mask that matches the human annotated mask. We trained three separate algorithms (5a/5b/5c), one for each of the predetermined aspect ratio image sizes (Aspect Ratio 1: 120 images, Aspect Ratio 2: 80 images, Aspect Ratio 3: 80 images), using the default hyperparameters as described[42]. Each image was normalized to the Z-score of each channel and augmented by left-right flip and rotation. Our datasets were split into 80% training and 20% test datasets. We trained the discriminator and the generator for successive epochs, alternatively, for up to 50,000 iterations. Learning rate was 2e-4, the optimizer was 'ADAM', the GAN2SEGMENTATION loss was 10:1 and the discriminator was dataset to the 'image-level'. We present the overall performance of Algorithms 5a/b/c in our test dataset as the sum of the Area Under the Receiver Operating Characteristic curve (AUC) and the Area Under the Curve for Precision and Recall Curve (PR-AUC). A sum of these two metrics of 2.00 suggests perfect segmentation, meaning that the mask generated by Algorithm 5 perfectly overlapped the ground-truth mask. We also present the Dice coefficient which represents the area of overlap divided by the total pixels between the predicted vessel mask and the traced vessel mask[43]. For the dice coefficient, we thresholded the probability map with the Otsu threshold[44] which is used to separate foreground pixels from background pixels.

Algorithm 6 was a modified Xception similar to, but trained separately from, Algorithm 4. Algorithm 6 took as input the same images as Algorithm 4 masked by the Algorithm 5-predicted masks. Due to the black-box nature of DNN algorithms, our Algorithm 5-6 sensitivity analysis also helped determine to what degree background elements in the image contributed to Algorithm 4's AI-stenosis prediction.

Algorithm 5 demonstrated excellent segmentation performance on the test dataset (average DICE coefficient of 0.79, AUC of 0.88, AUC-PR of 0.82 and a Sum of AUC and AUC-PR of 1.71). Algorithm 5 predicted coronary artery boundaries from cropped input images, trained by ground-truth masks. Algorithm 6 was trained using the same REPORT-stenosis labels as Algorithm 4. The Algorithm 6 predicted stenoses was strongly correlated with those from Algorithm 4 at the artery-level (r=0.70) in the test dataset. The average difference in predicted stenosis severity between Algorithm 4 and 6 was 12.4±10.9%, with 12.9±11.2% for right coronary arteries and 12.0±10.6% for left coronary arteries. These results provide evidence that Algorithm 4 performance does not substantially rely on image features outside of the coronary artery boundaries for predicting stenosis severity.

**Statistical analysis**

All analyses were performed using Python 2.7. Neural networks were trained using Keras v.2.24 and TensorFlow v.1.12. Final algorithms performance was reported in the Test Dataset.

Algorithm 1 and Algorithm 2 results are presented on the frame/image level. For each algorithm, we present class performance calculated using precision (i.e. positive predictive value), recall (sensitivity) and plot the performance using confusion matrices. We also derived the F1 score for each class, which is the harmonic mean between the precision and recall. This value ranges between 0 and 1 and is highest in algorithms that maximize both precision and recall of that class simultaneously.

To measure the performance of Algorithm 3a and 3b, we measured the IoU between the predicted coordinates and the actual coordinates on our test dataset. The IoU is the ratio between the area of overlap over the area of union between the predicted and annotated sets of coordinates[39]. The performance of algorithm 3 a/b is reported as the mean average precision (mAP), which represents the ratio of true positives over true and false positives at different thresholds of IoU, starting from IoU of 0.5, with steps of 0.05, to a maximum IoU of 0.95, for each class. We also present the mean average precision for algorithm 3a and algorithm 3b by calculating the proportion of correct class prediction with an IoU≥0.5 with our ground-truth labelling across all our classes in our test dataset.

Algorithm 4 is used to derive the average absolute error between the reported value (REPORT-stenosis) and the predicted value (AI-stenosis), at the artery-level, which was the primary aim of CathAI. This mirrors guideline-based standard clinical practice for stenosis estimation, by measuring stenosis in multiple orthogonal projections and reporting the maximal degree of stenosis narrowing[11]. Image-level AI-stenosis was averaged across a video to obtain video-level AI-stenosis. The mean of AI-stenosis predictions from all videos that visualized an artery segment in a study provided the artery-level AI-stenosis. We present Pearson correlation and Bland-Altman plots[45] to describe agreement between the REPORT-stenosis and AI-stenosis at the video-level and artery-level. Intra-class correlation (ICC) was used for interobserver reliability (between REPORT-stenosis and AI-stenosis). The reliability was classified as

slight (0.0-0.20), fair (0.21-0.40), moderate (0.41-0.60), substantial (0.61-0.80), or excellent (0.81-1.0).[21] Finally, we present the mean squared error between REPORT-stenosis and AI-stenosis at the video-level.

As a sensitivity analysis, we dichotomized the predicted and reported stenoses into two groups (≥70% and <70%). The 70% stenosis cutoff obtained by visual assessment defines an obstructive coronary stenosis[46]. We recalibrated Algorithm 4 percentage outputs by obtaining the AI-stenosis threshold that maximized the F1 score in our development dataset. In the Test dataset, we calculated AUC, sensitivity, specificity and diagnostic odds-ratio[47], at the frame level, video level and artery level, based on this threshold. Confidence intervals for performance metrics were derived by bootstrapping 80% of the test data over 1000 iterations to obtain 5th and 95th percentile values. We present the performance of the algorithm stratified by the left and right coronary arteries, by artery segment and by age group. We categorized AI-stenosis and REPORT-stenosis in concordant and discordant lesion groups based on the visual ≥70% cutoff. For discordant lesions we present their prevalence, stratified by coronary artery segment. For lesion/vessel level data, a mixed effects logistic regression model as used to account for within-subject correlation and for repeated angiograms.

For Algorithm 5, we trained a separate version of the algorithm that classified stenoses as ≥70% and <70% to derive LOVI saliency maps. To see the influence of the background elements in the AI-stenosis prediction, we also report the ICC, Pearson and the mean stenosis difference between our Algorithm 4a (taking vessels with the background as input) and Algorithm 5 (taking the segmented vessels, without their background)

**Data Availability**: The data used in this study are derived from clinical care and thus are not made publicly available due to data privacy concerns. Reasonable requests for collaboration using the data can be made from the authors, as feasible and permitted by the Regents of the University of California.

**Code Availability**: The code that supports this work is copyright of the Regents of the University of California and can be made available through license.

# Supplemental Materials

**Extended Figure 1.**

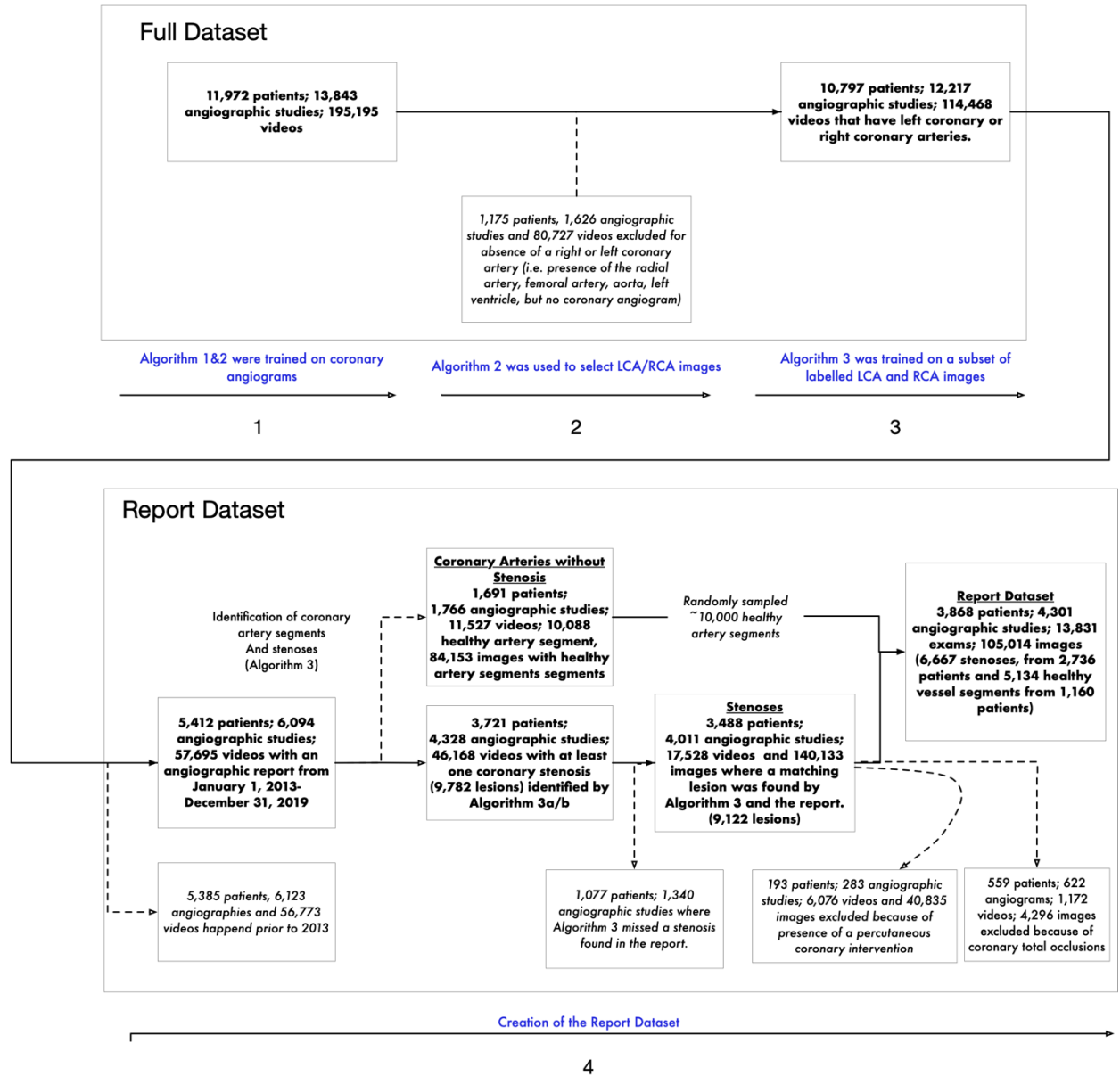

**Extended Figure 1. Datasets and patients used to develop CathAI algorithms.**
Detailed description of the Full Dataset and the Report Dataset. Step 1: We trained Algorithm 1 on the Full Dataset and Algorithm 2 on a subset of the Full Dataset to identify angiographic projections and anatomical structures, respectively. Step 2: We applied Algorithm 2 to the Full dataset to identify videos with left or right coronary arteries. Step 3: We selected left or right coronary artery videos and randomly selected a subsample to manually annotate for Algorithm 3 training. Step 4. To generate the Report Dataset, only angiographic studies associated with a digital procedural report were retained (those performed after 2013). We then cross matched the studies with a procedural report with stenosis identified by Algorithm 3. **Full arrows:** Creation of the study datasets; **Dashed arrows:** Excluded data. Abbreviations: LCA: Left Coronary Artery, RCA: Right Coronary Artery.

**Extended Figure 2.**

| Class | Precision | Recall | F1 Score | Frames |
|---|---|---|---|---|
| **RAO Cranial** | 0.88 | 0.93 | 0.90 | 37,638 |
| **AP Cranial** | 0.88 | 0.86 | 0.87 | 28,957 |
| **LAO Cranial** | 0.91 | 0.93 | 0.92 | 44,885 |
| **RAO Straight** | 0.91 | 0.86 | 0.88 | 39,563 |
| **AP** | 0.81 | 0.83 | 0.82 | 15,707 |
| **LAO Straight** | 0.94 | 0.91 | 0.93 | 57,028 |
| **RAO Caudal** | 0.84 | 0.86 | 0.85 | 19,552 |
| **AP Caudal** | 0.90 | 0.88 | 0.89 | 22,416 |
| **LAO Caudal** | 0.93 | 0.96 | 0.94 | 31,665 |
| **LAO Lateral** | 0.91 | 0.88 | 0.90 | 1570 |
| **RAO Lateral** | 0.73 | 0.66 | 0.70 | 485 |
| **Other** | 0.18 | 0.08 | 0.11 | 159 |
| **Overall weighted average** | **0.90** | **0.90** | **0.90** | **299,625** |

**Extended Figure 2. Cath AI classification performance of angiographic projection angle at the image-level in the test dataset (Algorithm 1).** Results are calculated on the hold-out Test Dataset for Algorithm 1. **Abbreviations:** RAO: Right Anterior Oblique; AP: Antero-posterior; LAO: Left Anterior Oblique.

**Extended Figure 3.**

a.

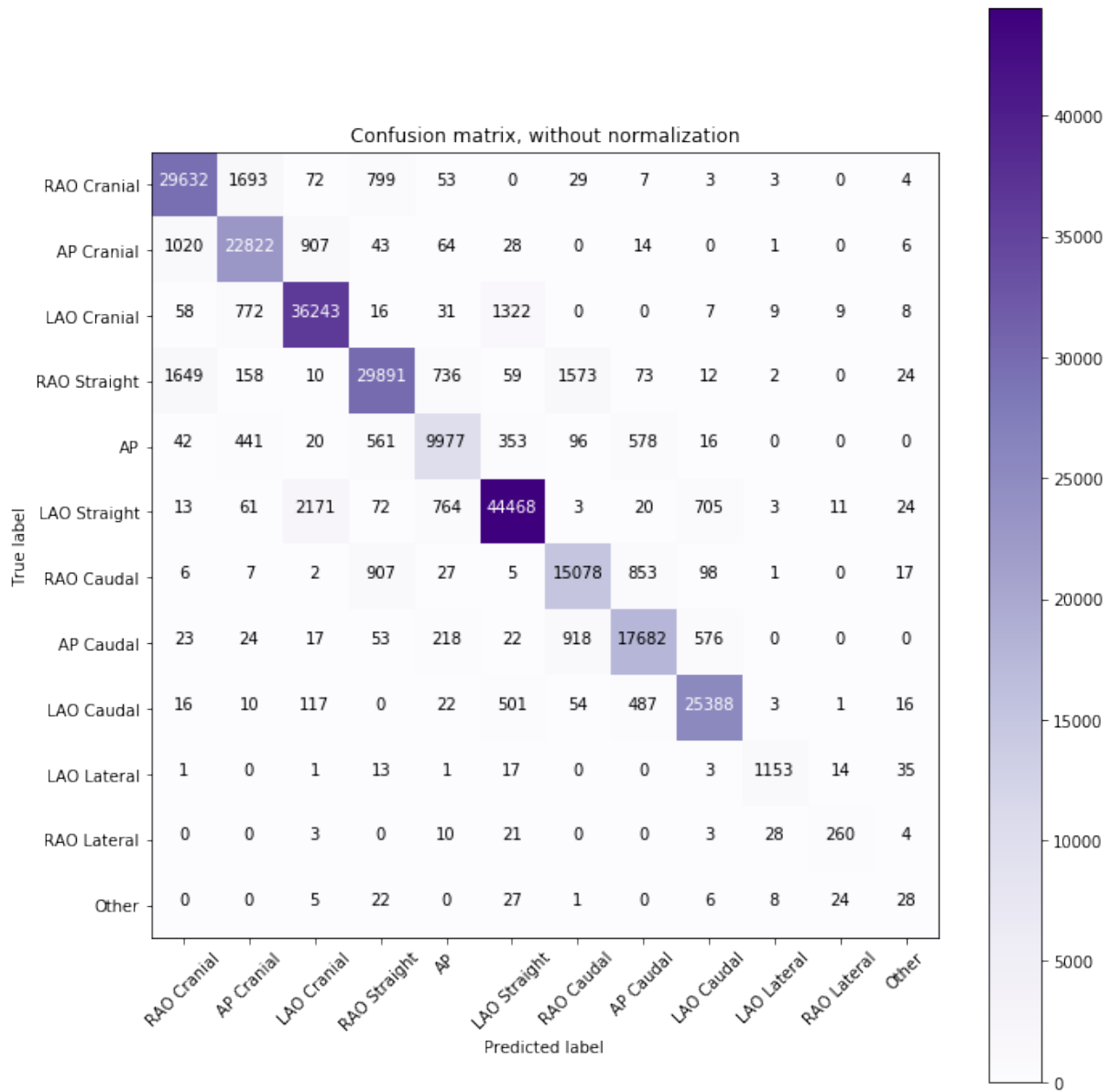

b.

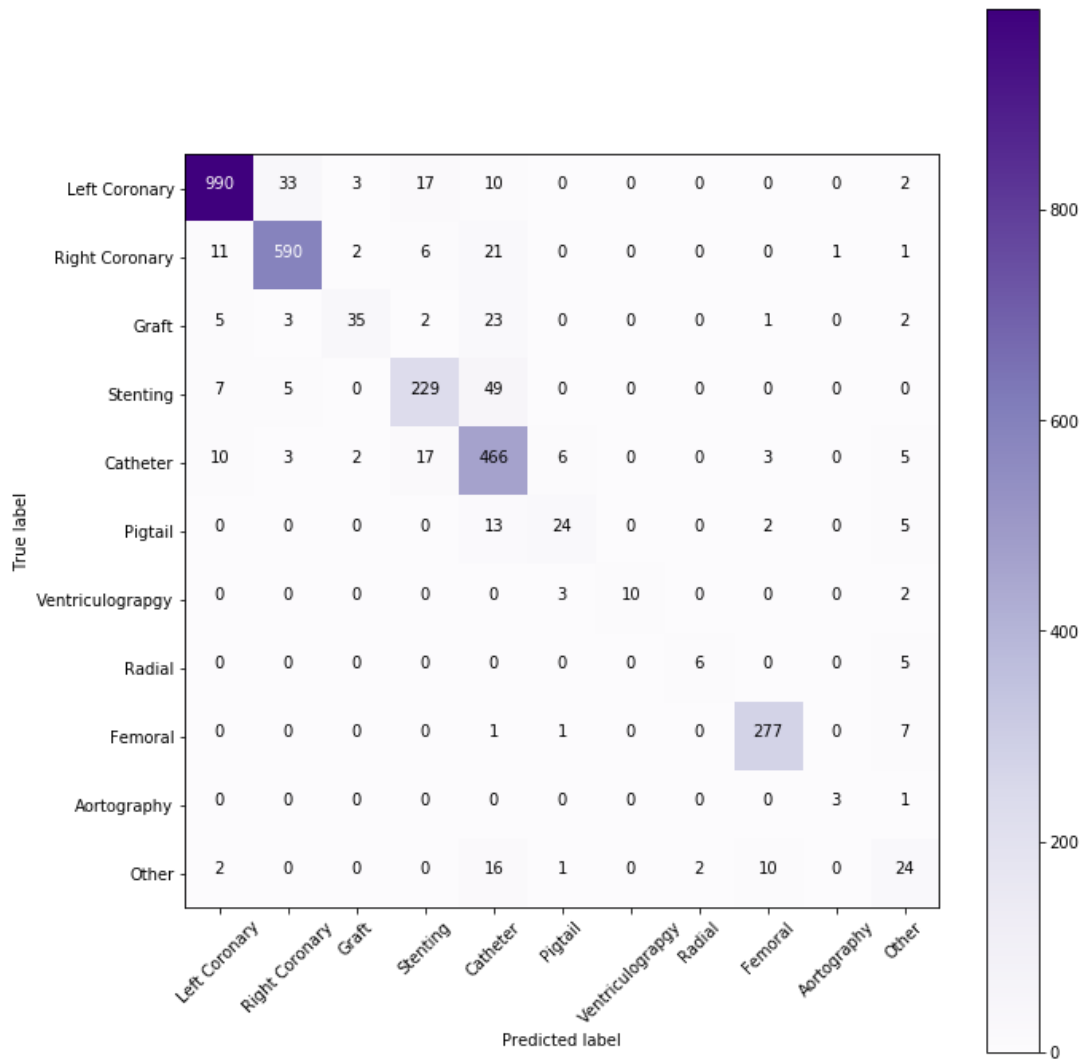

**Extended Figure 3. Confusion matrices for classification of angiographic projection angle and anatomic structure at the frame-level. a. Algorithm 1 - Confusion Matrix for classification of angiographic projections at the frame-level in the test dataset. b. Algorithm 2 - Confusion Matrix of the classification of the different anatomic structures, at the frame level in the test dataset.**
Abbreviations: RAO: Right Anterior Oblique; AP: Antero-posterior; LAO: Left Anterior Oblique.

**Extended Figure 4. CathAI performance for object localization (Algorithm 3).**

| Classes | Number of object labels in the Test Dataset (N) | Average Precision (%) |
|---|---|---|
| **Algorithm 3a: Left and right coronary artery (N = 234 images)** | | |
| **Left coronary artery segments** | **285** | **37.0** |
| Left main* | 48 | 49.8 |
| Proximal LAD* | 70 | 20.9 |
| Mid LAD* | 43 | 34.8 |
| Distal LAD* | 44 | 40.5 |
| Proximal/mid circumflex* | 53 | 41.0 |
| Distal circumflex* | 27 | 35.8 |
| **Right coronary artery segments** | **353** | **42.8** |
| Proximal RCA * | 81 | 50.2 |
| Mid RCA * | 70 | 41.8 |
| Distal RCA * | 82 | 29.8 |
| PDA * | 72 | 50.6 |
| PL * | 48 | 41.7 |
| **Other classes** | | |
| Stenosis | 93 | 13.7 |
| Stent/Balloon | 16 | 9.8 |
| Catheter | 208 | 76.1 |
| Guidewire | 8 | 73.2 |
| Sternotomy | 78 | 81.6 |
| Valve | 3 | 100.0 |
| Pacemaker | 19 | 64.9 |
| **Overall Weighted average** | **-** | **48.1** |

| Algorithm 3b: Right coronary artery in LAO straight projection (N = 45 images in the Test Dataset) | | |
|---|---|---|
| **Right coronary artery segments** | **118** | **54.5%** |
| Proximal RCA * | 29 | 40.2 |
| Mid RCA * | 24 | 83.2 |
| Distal RCA * | 28 | 55.4 |
| PDA * | 22 | 36.7 |
| PL * | 15 | 57.1 |
| **Other Classes** | | |
| Stenosis | 23 | 26.0 |
| Obstruction | 2 | 0.0 |
| Stent/Balloon | 2 | 0.0 |
| Catheter | 45 | 79.2 |
| Guidewire | 8 | 73.2 |
| Sternotomy | 12 | 92.6 |
| Valve | 3 | 66.7 |
| Pacemaker | 9 | 51.3 |
| **Overall Weighted Average†** | **-** | **58.1** |

Abbreviations: LAD: Left Anterior Descending; RCA: Right Coronary Artery; PDA: Posterior Descending Artery; PL: Posterolateral; Mid: Middle; mAP: Mean Average Precision.

*Different coronary artery sub-segments.

†: Average is weighted by the frequency of each class.

## Supplemental Tables

**Supplemental Table 1. Class Definitions of Angiographic Projection Angle Used for Algorithm 1**

| Class | Definition |
| --- | --- |
| RAO Cranial | -45° to -15° RAO; 15° to 45° Cranial |
| AP Cranial | -15° to 15° AP; 15° to 45° Cranial |
| LAO Cranial | 15° to 45° LAO; 15° to 45° Cranial |
| RAO Straight | -45° to -15° RAO; -15° to 15° AP |
| AP | -15° to 15° AP; -15° to 15° AP |
| RAO Caudal | -45° to -15° RAO; -45° to -15° Caudal |
| AP Caudal | -15° to 15° AP; -45° to -15° Caudal |
| LAO Caudal | 15° to 45° LAO; -45° to -15° Caudal |
| LAO Straight | 15° to 45° LAO; -15° to 15° AP |
| LAO Lateral | 70° to 110° LAO; -15° to 15° AP |
| RAO Lateral | -110° to -70° RAO; -15° to 15° AP |
| Other | Any angles not belonging to the previous definitions |

**Abbreviations:** RAO: Right Anterior Oblique; AP: Antero-posterior; LAO: Left Anterior Oblique.

**Supplemental Table 2. Classes of Primary Anatomic Structures Used for Algorithm 2**

| Class | Definition |
|---|---|
| **Left coronary artery** | Artery that arises from the aorta above the left cusp of the aortic valve |
| **Right coronary artery** | Artery that arises from the aorta above the right cusp of the aortic valve |
| **Bypass Graft** | Venous graft, internal mammary graft or radial graft |
| **Catheter** | Any guiding catheter or diagnostic catheter without any other underlying structure |
| **Pigtail Catheter** | Pigtail catheter without any other underlying structure |
| **Left ventricle** | Ventricle, as delimited during ventriculography |
| **Aorta** | Ascending aorta, the arch or descending aorta, as delimited during aortography |
| **Radial Artery** | Major artery in the forearm |
| **Femoral Artery** | Either the superficial, deep or common femoral artery |
| **Other** | Any images not belonging to the other classes (for example, kidneys, pacemaker, etc) |

**Supplemental Table 3. Definition of Objects Used for Algorithm 3**

| Class | Definition |
|---|---|
| **Coronary segment** | |
| Proximal RCA* | From ostium to one half the distance to the acute margin of the heart. |
| Middle RCA* | From end of first segment to acute margin of heart. |
| Distal RCA* | From the acute margin of the heart to the origin of the posterior descending artery. |
| Posterior descending artery* | Artery running the posterior interventricular groove. |
| Posterolateral artery* | Posterolateral branch originating from the distal coronary artery distal to the crux. If left posterolateral, it was chosen as the artery running to the posterolateral surface of the left ventricle. |
| Left main artery* | From the ostium of the LCA through bifurcation into left anterior descending and left circumflex branches. |
| Proximal LAD* | Vessel between left main and proximal to and including the first septal |
| Middle LAD* | LAD immediately distal to the origin of first septal branch and extending to the point where the LAD forms an angle (RAO projection). If angle is not identifiable, this segment ends at one half the distance form the first septal and the apex of the heart |
| Distal LAD* | Terminal portion of LAD, beginning at the end of previous segment and extending to or beyond the apex. |
| Proximal LCX* | Main stem of circumflex from its origin of left main to and including origin of first obtuse marginal branch. |
| Distal LCX* | The stem of the circumflex distal to the origin of the most distal obtuse marginal branch and running along the posterior left atrioventricular grooves. Caliber may be small or artery absent. |
| **Other classes** | |
| Valve | Presence of a mechanical valve, annuloplasty or valvular calcifications |
| Catheter | Presence of a catheter, such as a diagnostic catheter, pigtail or guiding catheter |
| Sternotomy | Presence of sternotomy wires |
| Stent | Stent landmarks on a guidewire or in a vessel |
| Pacemaker | Presence of a pacemaker or pacemaker lead |
| Guidewire | Presence of a guide wire |
| Stenosis | Any visible stenosis that is also described in the cath report. |
| Obstruction | 100% obstruction of an artery, either by thrombus or chronically occluded. Defined by a blunt stump at the end to a vessel or by the 'absence' of contrast in between two healthy vessel segments with bridging collaterals. |

**Abbreviations:** RCA: Right Coronary Artery; LAD: Left Anterior Descending Artery; LCX: Left Circumflex.

*These coronary vessel segments follow the SYNTAX score definition.[20]

**Supplemental Table 4. Heuristic Exclusion of Coronary Artery Segments by Angiographic Projection**

| Projection | Excluded segments | |
| --- | --- | --- |
| | Right Coronary Artery | Left Coronary Artery |
| **RAO Cranial** | None | Proximal LCx, Distal LCx |
| **AP Cranial** | None | Proximal LCx, Distal LCx |
| **LAO Cranial** | None | Proximal LCx, Distal LCx |
| **RAO Straight** | Proximal RCA | Proximal LAD |
| **AP** | None | Mid LAD, Distal LAD, Distal LCx |
| **RAO Caudal** | None | Mid LAD, Distal LAD |
| **AP Caudal** | None | Distal LAD |
| **LAO Caudal** | None | Mid LAD, Distal LAD |
| **LAO Straight** | None | Proximal LCx, Distal LCx |
| **LAO Lateral** | None | None |
| **RAO Lateral** | None | None |
| **Other** | None | None |